\documentclass[10pt,twocolumn,letterpaper]{article}

\usepackage{cvpr}              

\usepackage{graphicx}
\usepackage{amsmath}
\usepackage{amssymb}
\usepackage{booktabs}
\usepackage{bm}

\usepackage[pagebackref,breaklinks,colorlinks]{hyperref}

\usepackage[capitalize]{cleveref}
\crefname{section}{Sec.}{Secs.}
\Crefname{section}{Section}{Sections}
\Crefname{table}{Table}{Tables}
\crefname{table}{Tab.}{Tabs.}

\usepackage{xcolor}

\def\cvprPaperID{8116} 
\def\confName{CVPR}
\def\confYear{2023}

\begin{document}

\title{AgileGAN3D: Few-Shot 3D Portrait Stylization by Augmented Transfer Learning
}


\author{Guoxian Song$^1$ \quad Hongyi Xu$^1$ \quad Jing Liu$^1$ \quad Tiancheng Zhi$^1$ \quad Yichun Shi$^1$ \\  Jianfeng Zhang$^{1,2}$ \quad Zihang Jiang$^{1,2}$ \quad Jiashi Feng$^1$ \quad Shen Sang$^1$ \quad Linjie Luo$^1$\\
$^1$ByteDance Inc \quad\quad $^2$National University of Singapore\\
}

\maketitle

\newcommand{\bb}[1]{\boldsymbol{\mathbf{#1}}}
\begin{abstract}

While substantial progresses have been made in automated 2D portrait stylization, admirable 3D portrait stylization from a single user photo remains to be an unresolved challenge. One primary obstacle here is the lack of high quality stylized 3D training data. In this paper, we propose a novel framework \emph{AgileGAN3D} that can produce 3D artistically appealing and personalized portraits with detailed geometry. New stylization can be obtained with just a few (around 20) unpaired 2D exemplars. We achieve this by first leveraging existing 2D stylization capabilities, \emph{style prior creation}, to produce a large amount of augmented 2D style exemplars. These augmented exemplars are generated with accurate camera pose labels, as well as paired real face images, which prove to be critical for the downstream 3D stylization task. Capitalizing on the recent advancement of 3D-aware GAN models, we perform \emph{guided transfer learning} on a pretrained 3D GAN generator to produce multi-view-consistent stylized renderings. In order to achieve 3D GAN inversion that can preserve subject's identity well, we incorporate \emph{multi-view consistency loss} in the training of our encoder. Our pipeline demonstrates strong capability in turning user photos into a diverse range of 3D artistic portraits. Both qualitative results and quantitative evaluations have been conducted to show the superior performance of our method. Code and pretrained models will be released for reproduction purpose.

\end{abstract}


\section{Introduction}
\begin{figure}[hbt!]
\centering
\includegraphics[width=1\linewidth]{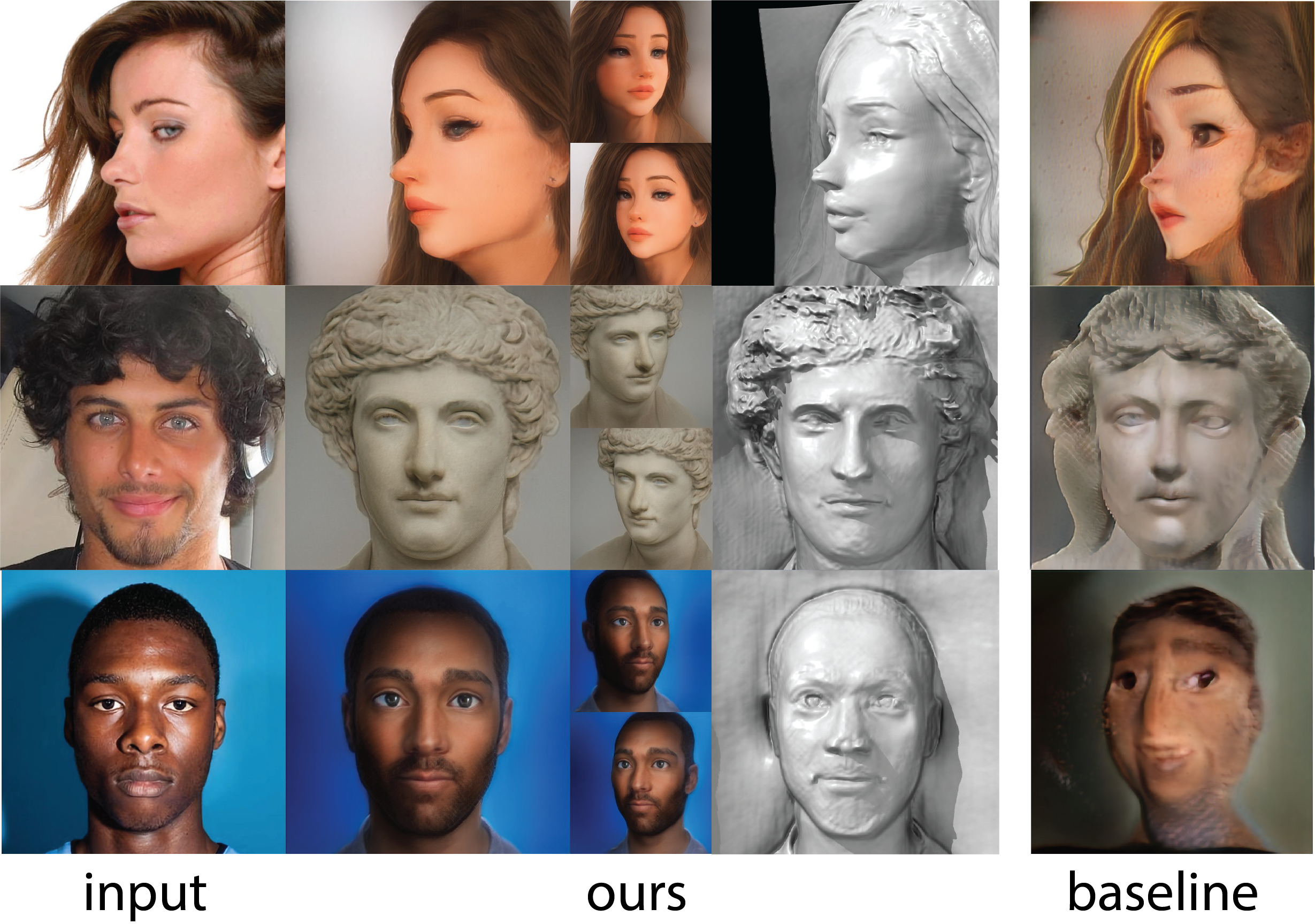}
\centering
\caption{Our AgileGAN3D enables  3D stylized portraits creation from a single input image. A new 3D style can be obtained with only a few unpaired 2D style exemplars ($\sim$ 20). Compared to the baseline method (directly fine-tune the 3D GAN model~\cite{chan2022efficient}), our approach produces high-quality, multi-view-consistent renderings of the portrait, with detailed in-style geometry. }
\label{fig:teaser}
\end{figure}


Portrait painting as an art form goes back to prehistoric times, and has been primarily serving the rich and powerful. Fast forward with the technology advancement, people nowadays can enjoy a high fidelity digital portrait within seconds, and even capturing one's detailed facial 3D geometry\cite{Chai:2015:HQH,3DPortraitfrom}.
Driven by the creative nature of human beings, people are no longer satisfied with simply a faithful depiction of their appearance. Portraiture has evolved into more expressive interpretations with a plethora of styles, such as abstract art, cubism and cartoon. However, most previous works are limited to stylized portraits in 2D image space\cite{pinkney2020resolution, Song:2021:AgileGAN,yang2022Pastiche,chong2021jojogan}. Automatically creating 3D stylized portrait with detailed geometry using just a single selfie as input is still an open problem. To the best of our knowledge, we are the first work that can automatically create 3D stylized portrait with detailed geometry, using just a single selfie as input. The result 3D portrait can be adapted into a wide range of artistic styles, as long as a few 2D style exemplars are provided (see Fig.\ref{fig:teaser}). Such new format enables a lot of applications like 3D printed postcards, dynamic profile pictures (by changing viewing angles, or lighting directions), as well as personalized 3D contents in augmented and virtual reality worlds.

The core challenge that prevents us from creating visually appealing, personalized 3D portraits is rooted in the shortage of high quality 3D data. A traditional approach to create customized 3D portraits for users is by assembling a 3D avatar system with tons of graphics assets (e.g. Zepeto\footnote{https://zepeto.me/}, ReadyPlayer\footnote{https://readyplayer.me/}, and \cite{Sang2022AgileAvatar}). However, it's almost impossible to capture all the diversities in real world appearances given only a few hundred assets and a base morphable 3D face model. Therefore such approach usually generates less personalized results.

The latest generative models such as StyleGAN2~\cite{karras2020analyzing}, latent diffusion models~\cite{Rombach_2022_CVPR} are very powerful architectures in producing highly diversified imageries, largely credited to the huge data sources that these models have been trained on. Fine-tuning a generative model to produce highly personalized portraits thus becomes possible, though still primarily in 2D image space\cite{pinkney2020resolution, Song:2021:AgileGAN,yang2022Pastiche,chong2021jojogan}. Lifting arbitrary stylized portraits from 2D into 3D space remains to be an unsolved problem, partly due to the lack of 3D prior knowledge for target artistic pictures. Promising early results have emerged for general 3D objects~\cite{poole2022dreamfusion,jain2021dreamfields}, but none of them can produce desirable quality for 3D portraits yet. 

Recent rapid progress in geometry-aware GANs \cite{chan2021pi,niemeyer2021giraffe,chan2022efficient,or2021stylesdf,deng2021gram,epigraf,zhou2021CIPS3D} inspired our work. Particularly EG3D ~\cite{chan2022efficient} demonstrated strikingly realistic 3D face synthesis capability by using only unstructured 2D photos. Its tri-plane structure serves as an efficient representation for 3D content generation, combined with a neural volumetric renderer, making multi-view-consistent, photo-realistic image synthesis possible. However, even with such a powerful 3D generator, there are still a few obstacles ahead before we can have a usable 3D stylized portrait generator. First, even in 2D image format, it is nontrivial to collect a large number of diverse portraits in a consistent style, let alone obtaining robust camera pose estimations from artistic portraits, which is a critical step in training 3D GANs. Secondly, in order to personalize user photos into 3D stylized portraits, a reliable 3D GAN inversion method is required. The inversion module has to work in a way that balances the reconstruction fidelity and the stylization quality.

To tackle the aforementioned challenges, we propose \emph{AgileGAN3D}, a novel \emph{augmented transfer learning} framework for generating high quality stylized 3D portraits, using only a few 2D style exemplars ($\sim$20).
With extensive experiments, we observe the successful transfer learning for 3D GANs relies on adequate style visual supervisions with well estimated camera labels. 
To address the shortage of stylized training data issue, we started with \emph{style prior creation}, that leverages the existing 2D portrait stylization capabilities. Specifically, we trained a 2D portrait stylization module following AgileGAN\cite{Song:2021:AgileGAN}, to first obtain a large number of stylized portraits using real face photos as inputs. 
Some extra benefits of this way of generating 2D style exemplars are that we naturally obtain: 1) pair data between stylized faces and real faces; 2) fairly accurate head pose estimations of generated stylized portraits (by reusing the poses estimated from the corresponding real faces). 
Both of these benefits are incorporated into our \emph{guided transfer learning} step with a reconstruction loss, that helps improve the 3D stylization for out-of-domain samples. Equipped with a transfer-learned style generator, we further introduce an 3D GAN encoder that embeds a real image into an enlarged latent space for better identity-preserved 3D stylization.  A cycle consistency loss is proposed in the 3D GAN encoder training to further improve the multi-view reconstruction fidelity.
To best of our knowledge, we are the first paper to propose generative NeRF based 3D stylized portrait creation using only a limited number of 2D style exemplars.

To summarize the contributions of our work:

\begin{itemize}
\vspace{-0.05in}
\item A novel pipeline for creating 3D stylized portraits with detailed geometry, given only a single user photo as input. New stylization can be achieved with only a few unpaired 2D style exemplars (around 20). 
\vspace{-0.05in}
\item A simple yet efficient way to fine-tune 3D GAN, first with \emph{style prior creation} to improve data diversity, combined with \emph{guided transfer learning} to increase the stylization domain coverage;
\vspace{-0.05in}
\item A 3D GAN encoder that inverts real face images into corresponding latent space, trained with cycle consistent loss to improve identity preservation, while achieving high stylization quality.
\end{itemize}

\section{Related Work}

\paragraph{Face Stylization}
Stylizing facial images in an artistic manner has been explored in the context of non-photorealistic rendering. Early approaches relied on low level histogram matching using linear filters~\cite{10.1145/218380.218446}. Neural style transfer~\cite{7780634}, by matching feature statistics in convolutional layers, led to early exciting results via deep learning. However, they usually fail on styles involving significant geometric deformation of facial features, such as cartoonization. For more general cross-domain stylization, Toonify~\cite{pinkney2020resolution} proposed a GAN interpolation framework for controllable cross-domain image synthesis for cartoonization.
A following method AgileGAN~\cite{Song:2021:AgileGAN} proposed VAE inversion to enhance distribution consistency in the latent space, leading to fewer artifacts and better results for real input images. Besides, Huang \etal ~\cite{huang2021unsupervised} achieves multi-domain stylization via a layer swapping technique. Recent exemplar-based approaches~\cite{yang2022Pastiche,chong2021jojogan,li2021anigan} enable one-shot portrait stylization given a single style exemplar. There are also several 2D stylization works for video generations~\cite{yang2022Vtoonify,back2022webtoonme}. In contrast, our proposed \emph{AgileGAN3D} produces highly detailed 3D stylized portraits using the same amount of input as used in 2D stylization methods.



\paragraph{Geometry-Aware GANs}
Generative adversarial networks~\cite{goodfellow2014generative,karras2019style,Karras2020stylegan2} have been used to synthesize images ideally matching the training dataset distribution via adversarial learning.
Built on the success of 2D GANs, recent works have extended to multiview consistent image synthesis with unsupervised learning from unstructured single-view images. The key idea is to combine differential rendering with 3D representations, such as meshes~\cite{Szabo:2019,Liao2020CVPR,gao2022get3d}, point clouds~\cite{li2019pu}, voxels~\cite{wu2016learning,hologan,nguyen2020blockgan}, and recently implicit neural representation~\cite{chan2021pi,niemeyer2021giraffe,chan2022efficient,or2021stylesdf,deng2021gram,epigraf,zhou2021CIPS3D}. Especially the Neural Radiance Fields (NeRF)\cite{gu2021stylenerf,or2021stylesdf} representations, which have proven to generate high-fidelity results in novel view synthesis, are introduced to 3D-aware generative models. They typically use StyleGAN2~\cite{karras2020analyzing} as the backbone to generate intermediate features for MLP to query and perform neural volumetric rendering for image synthesis. Recently, EG3D\cite{chan2022efficient} uses an efficient triplane-based neural radiance field, combined with CNN-based upsampling and a pose-aware dual discriminator to improve synthesis quality and multi-view consistency. Our work further extends the success of existing 3D GAN models to generate stylized results with only a few unpaired 2D stylized exemplars.

\paragraph{GAN Inversion}
Given an input image, GAN inversion addresses the complementary problem of finding the most accurate latent code to reconstruct that image. Existing approaches roughly fall into three categories: optimization-based, learning-based and hybrid GAN inversion. Optimization-based approaches \cite{karras2019stylebased,Image2StyleGAN,10.1145/3414685.3417803} directly optimize the latent code to minimize the pixel-wise reconstruction loss for a single input instance. Learning-based approaches~\cite{zhu2016generative} train a deterministic model by minimizing the difference between the input and synthesized images. Some works combine these ideas, e.g. learning an encoder that produces a good initialization for subsequent optimization \cite{bau2019seeing}. In addition to image reconstruction, some methods also use inversion when undertaking image manipulation. For example, Zhu \etal~\cite{zhu2020indomain} introduced a hybrid method to encode images into a semantic manipulable domain for image editing. Richardson \etal~\cite{richardson2021encoding} presented the generic Pixel2Style2Pixel (pSp) encoder to embed image into StyleGAN $W^+$ space, based on a dedicated identity loss. 
To balance reconstruction and editing ability for inversion, E4E encoder~\cite{tov2021designing} later uses a progressive training strategy to stimulate $W$ space for $W^+$ space. ReStyle~\cite{alaluf2021restyle} trains a residual based encoder with iterative refinement. Besides training encoder, Pivotal Tuning Inversion\cite{roich2021pivotal} also fine-tunes the generator parameters each time for recovering image details that cannot be encoded in the latent space to improve reconstruction. Wang \etal ~\cite{wang2022high} proposed an approach to achieve high fidelity inversion without inference time optimization. Recent works~\cite{alaluf2022hyperstyle,dinh2022hyperinverter} employ hyper networks~\cite{ha2016hypernetworks} to improve StyleGAN inversion. However, directly adopting 2D GAN encoders for 3D GAN models won't work well, as it is not taking account of the multi-view generation aspect of a 3D GAN model. In our work, we introduce a \emph{multi-view cycle consistent loss} to improve our 3D GAN encoder's capability to preserve subject identities while balancing the stylization quality.


\section{Method}

\begin{figure*}[hbt!]
	\centering
	\includegraphics[width=1.0\linewidth]{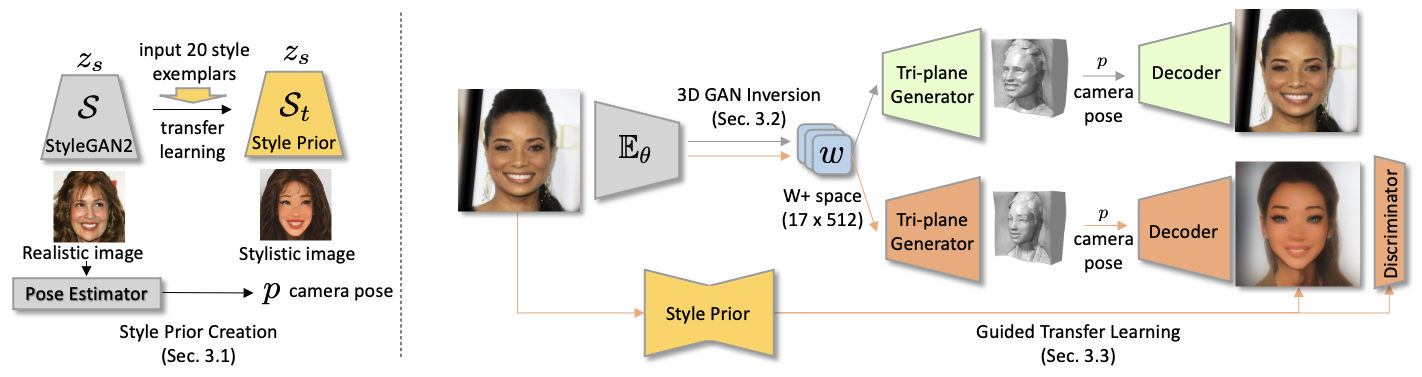}
	\centering
	\caption{Pipeline overview. Our 3D stylization pipeline consists of three stages, style prior creation, 3D GAN inversion and guided style transfer learning, with different data training flows indicated in different colors. Specifically given a few 2D style exemplars, we create a 2D style prior (left, yellow) that augments stylistic training samples with well-estimated camera labels from real images. We then perform transfer learning of 3D-aware image generator to target styles using augmented labeled style samples (orange), under the paired guidance of 2D stylization. Additionally we train an  encoder for 3D GAN image inversion (green) into a corresponding latent code in $W^+$ space, from which we can turn into an identity-preserved 3D style portraits.}  
	\label{fig:overview}
\end{figure*}

As shown in Fig.~\ref{fig:overview}, we first use \emph{style prior creation} to augment the limited 2D style exemplars, in order to supply downstream 3D GAN transfer learning with sufficient training data with well-estimated camera labels (Section~\ref{sec:spc}). Then we train an encoder to map input images into 3D GAN latent space ($W^+$), which well preserves facial identities with a \emph{multi-view cycle consistent loss} (Section~\ref{sec:inversion}). To further improve the stylization quality, we add \emph{guided transfer learning} that removes out-of-domain stylization artifacts(Section~\ref{sec:transfer}). 

\paragraph{3D-Aware Image Generation.}
\label{sec:EG3D}
For multi-view consistent image generation, we build our pipeline on top of a state-of-the-art geometry-aware 3D GAN model named EG3D~\cite{chan2022efficient}. To synthesize an image, the 3D generator $\mathcal{G}_{\phi}$ takes two variables: a latent code $z$, from a standard Gaussian distribution, that determines the geometry and appearance of a subject; a conditional camera pose label $\hat{p}$ added to the latent code. $z$ and $\hat{p}$ are passed through a multi-layer perceptron (MLP) mapping network to obtain a $w$ code, which is duplicated multiple times to 
modulate the synthesis convolution layers that produce tri-plane features. These features are sampled into a neural radiance field at the desired camera angle $p$ and accumulated to generate a raw feature image via volumetric rendering. Finally, the raw feature images are up-sampled by a super resolution module to synthesize the final RGB images. A camera-conditioned dual discriminator $D$ is used to examine the image fidelity in adversarial training, while ensuring multi-view consistency.

\subsection{Style Prior Creation}
\label{sec:spc}

\begin{figure}
    \centering
    \includegraphics[width=1.0\linewidth]{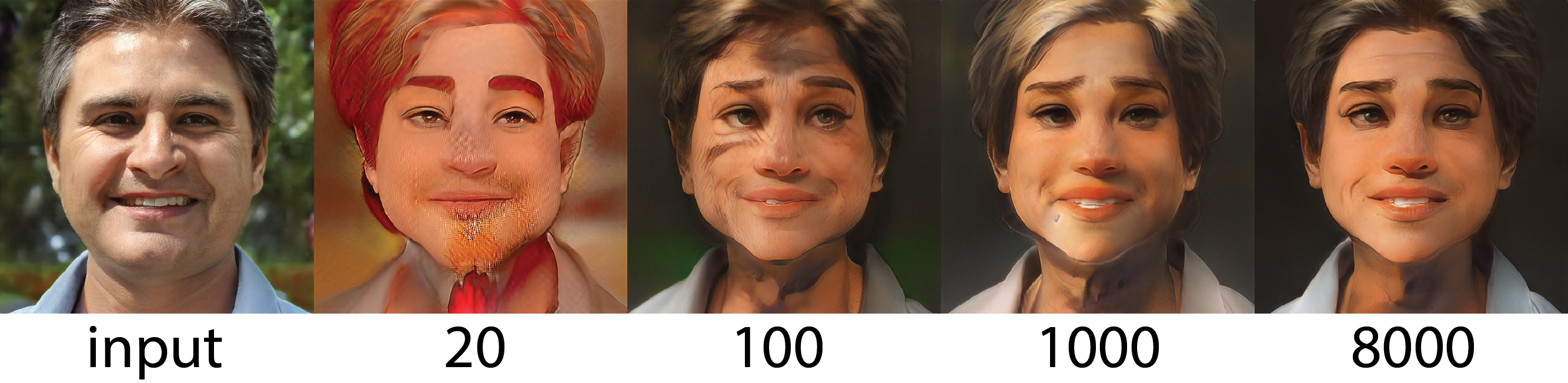}
    \centering
    \caption[Caption for LOF]{The evolution of transfer learning results using different number of training samples by our 2D style prior. Better visual quality is achieved with more training style exemplars. }
    \label{fig:alba_volume}
\end{figure}

Unlike the 2D GAN-based stylization tasks, the few-shot transfer learning on 3D GAN model is less well studied. A straightforward attempt to 3D portrait stylization will be through fine-tuning a pretrained 3D generator $\mathcal{G}_{\phi}$ directly with the few shot samples, e.g., 20 stylized exemplars. However, plain transfer learning generates poorly in both perceptual quality and user similarity. 
We suspect the problem is rooted in two aspects:
insufficient style exemplars due to the more complicated nature of a 3D GAN architecture and using inaccurate camera pose estimation from style exemplars. 





To mitigate the above problems, instead of directly using given stylized exemplars to fine-tune the 3D generator, we create a style prior based on 2D GAN to guide the transfer learning, which are less complicated and does not need camera pose.
Here we leverage the capability of the state-of-the-art 2D stylization methods for \emph{style prior creation}. Inspired by recent 2D stylization works~\cite{pinkney2020resolution,Song:2021:AgileGAN}, we perform transfer learning on top of the original StyleGAN2 generator $\mathcal{S}$ trained on FFHQ dataset~\cite{karras2019style}, with the few shot style exemplars. We denote the fine-tuned styled generator as $\mathcal{S}_{t}$. This gives us the capability to turn widely-accessible real portrait images into augmented 2D style samples. Moreover, this augmentation approach naturally offers pairs of stylized and real images, from the latter of which we can obtain accurate camera pose labels with off-the-shelf pose estimator such as \cite{3DPortraitfrom}.

\paragraph{Transfer Learning Loss}
 By sampling from prior latent space, we can get infinite high-quality diverse stylized images for transfer learning on 3D GANs. We use an adversarial loss to fine-tune the pre-trained 3D-aware generator $\mathcal{G}$ with respect to its parameter $\phi$ as well as its dual discriminator $D$, that matches the distribution of the translated images to the style prior distribution:
\begin{equation}\label{eq:loss_decoder_adv}
\begin{split}
    \mathcal{L}_{prior} = &
      \mathbb{E}_{z_s \sim N(0,I) }[min(0, -1+D(\mathcal{S}_t(z_s), p))] + \\
    & \mathbb{E}_{z \sim N(0,I)} [min(0, -1-D(\mathcal{G}_{\phi}(z, p), p))]
\end{split}
\end{equation}
where the latent code $z_s$ and $z$ are from StyleGAN2 and EG3D latent space respectively. 
We also apply regularization terms for stable fine-tuning. For discriminators, we use R$_1$ path regularization.

Our style prior leads to significant quality improvements, as shown in Fig.~\ref{fig:alba_volume}. We believe this is a requirement imposed by the NeRF module inside the 3D GAN pipeline. Typically, visual observations from a wide camera distribution are necessary for NeRF to reason the underlying 3D scene geometry and its corresponding appearance. Thus fine-tuning the cross-domain 3D generation of neural radiance features requires much more style samples than prior 2D GAN-based approaches.


Another point is the camera pose. For certain artistic styles, direct camera pose estimations from the input style exemplars might not be very accurate, which also affects 3D stylization. Different from 2D StyleGANs that directly up-sample feature maps into images via several convolution layers, 3D GAN synthesizes images by first accumulating neural radiance features via volume rendering to a feature map and then rely on super resolution to obtain the final image. Both the volume rendering and dual discriminator require reliable estimation of camera parameters, which are not easy to accurately obtained from non-realistic style examples. 


\subsection{Encoder Inversion with $W^+$}
\label{sec:inversion}

\begin{figure}[t]
    \centering
    \includegraphics[width=1\linewidth]{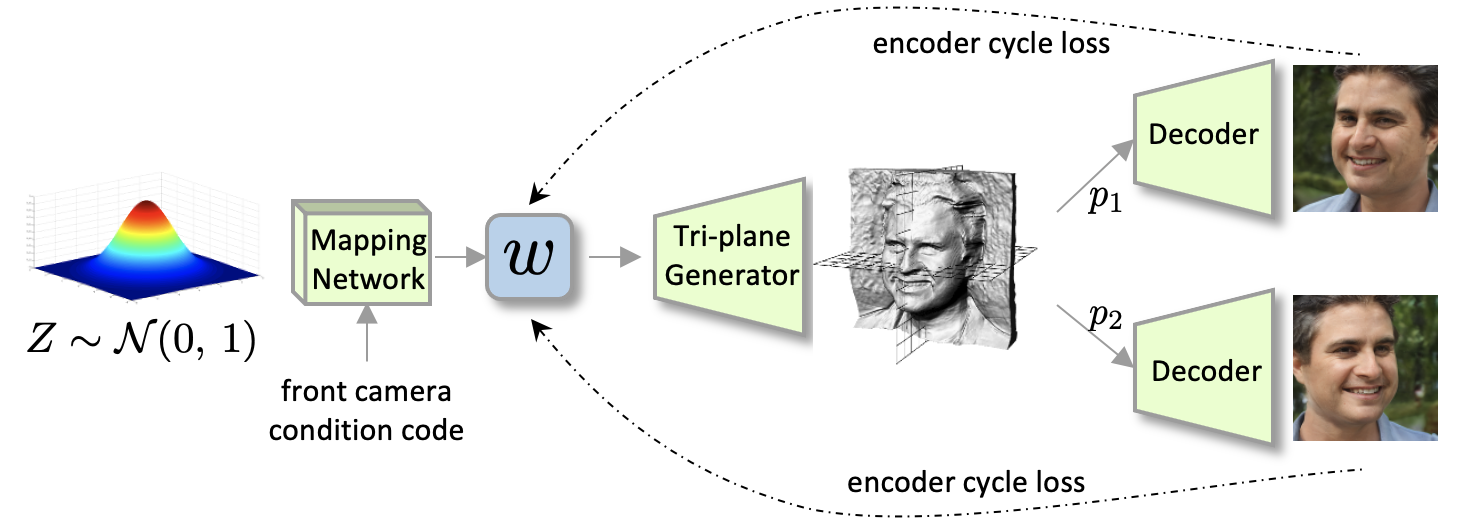}
    \centering
    \caption{Illustration of multiview cycle consistency loss. Sampling from the Gaussian-distributed latent space, we synthesize facial images at different random poses, from which we minimize the difference of our encoded latent codes with the input.}
    \label{fig:encoder}
\end{figure}

Assisted with the style prior creation step, we are able to achieve desirable 3D GAN stylization quality. One remaining challenge to complete the pipeline is the capability to invert a real face image into the latent space for 3D stylization.
\paragraph{Embedding Space and encoding}
 The pre-trained EG3D model~\cite{chan2022efficient} is equipped with two latent spaces: the original latent space $Z$ under a Gaussian distribution, and a less entangled $W$ space via a mapping network from $z$ with a conditional camera pose label $\hat{p}$. In spite of the camera swapping strategy, we observe the tri-plane generation is not fully decoupled from the pose, inducing varying geometry and appearance along with the change of $\hat{p}$. Therefore we use $W$ space for our image embedding, and augment it to $W^+$ space that significantly increases the model expressiveness. In contrast to modulating the convolutional kernels with a same  code $w$, $W^+$ produces a different $w$ latent code for each layer, allowing for individual attribute control. In our case, a code $w^+$ in $W^+$ space has a dimension of $17 \times 512$, which can be represented as 17 $w_i$ codes, where the $w_0$,...,$w_{13}$ codes are for tri-plane generation, and $w_{14},...,w_{16}$ are used in super resolution module. 

 

For fast inference, we train an encoder for image inversion, with the expectation of preserving user features to the largest extent. We design our encoder based on the architecture used in StyleGAN2 encoder, E4E~\cite{tov2021designing}, but fully exploit the unique proprieties of 3D GAN in generating view-consistent contents. In particular, we utilize the hierarchy of a pyramid network to capture different levels of detail from different layers. The input image resized at 256$\times$256 resolution is passed through a headless pyramid network $\mathcal{E}_\theta$ to produce three levels of feature maps at different sizes. Each level's feature map then goes through a separate sub-encoder block to produce the $W^+$ style code. 

\paragraph{Encoder Training Loss}
Even though our chosen $W^+$ space offers a large degree of freedom and expressiveness in representing real human faces, straightforward encoding without regularization can easily lead to out-of-domain issues, where the synthesized images present undesired artifacts like blurriness and noise. To prevent the encoder from over-drifting from the representation domain of $\mathcal{G_{\phi}},$ we introduce a \emph{multi-view cycle consistent loss}, as shown in Fig~\ref{fig:encoder}. The core idea is that the encoder should reproduce the latent code from a synthesized image conditioned on $w$ but rendered from arbitrary views. 
Specifically, a collection of latent codes randomly sampled under the standard Gaussian distribution, together with a fixed frontal camera pose, are fed into the mapping network and obtain $w$ samples. Note that these in-domain $w$ samples are complied with the original distribution of EG3D and likely to synthesize high-quality images without artifacts, and are a special form in $W^+$ space as well. Essentially training the encoder with these in-domain samples prevents the output $w^+$ codes from drifting far-away from the $W$ space. We synthesize the images with $N$ random camera poses $p_1, p_2,..$ from training dataset camera distribution and supervise the training of $\mathcal{E_\theta}$ with ground-truth $w^+$ labels.

\begin{equation} 
\label{eq:loss_cycle}
 \begin{split}
\mathcal{L}_{cyc} =\sum_{i=1}^{N}\mathcal{L}_{2}(w^+,\mathcal{E}_\theta(\mathcal{G_{\phi}}(w,p_i))).
 \end{split}
\end{equation}

In addition to the \emph{multi-view cycle consistent loss}, our encoder is at the same time trained with reconstruction losses and regularization loss in a weighted combination manner, while freezing the EG3D generator weights.


Let $x$ be the input image, passed through an encoder and decoder to yield $\hat{x} = \mathcal{G}_{\phi}(\mathcal{E}_\theta (x) )$
\begin{equation}
\label{eq:loss_rec}
\begin{split}
    \mathcal{L}_{rec} = \mathcal{L}_{2}(x, \hat{x}) + \mathcal{L}_{lpips}(x, \hat{x}) + \mathcal{L}_{arc}(x, \hat{x})
 \end{split}
\end{equation}
The $\mathcal{L}_{2}, \mathcal{L}_{lpips}, \mathcal{L}_{arc}$ respectively measure the pixel-level, perceptual-level similarities \cite{zhang2018perceptual} and facial recognition-level similarity differences. $\mathcal{L}_{arc}$ is based on the cosine similarity between intermediate features extracted from a pre-trained ArcFace recognition network~\cite{deng2018arcface}, evaluating the identity similarity. A regularization term is further introduced to reduce the divergence of $w^+$ code to mimic the origin $W$ space for the best of image quality, 
\begin{equation}\label{eq:loss_reg}
\mathcal{L}_{reg}   =  || \textrm{div}(\mathcal{E}_\theta (x)) ||_2.
\end{equation}



\subsection{Guided Transfer Learning}
\label{sec:transfer}

\begin{figure}[h]
    \centering
    \includegraphics[width=1.0\linewidth]{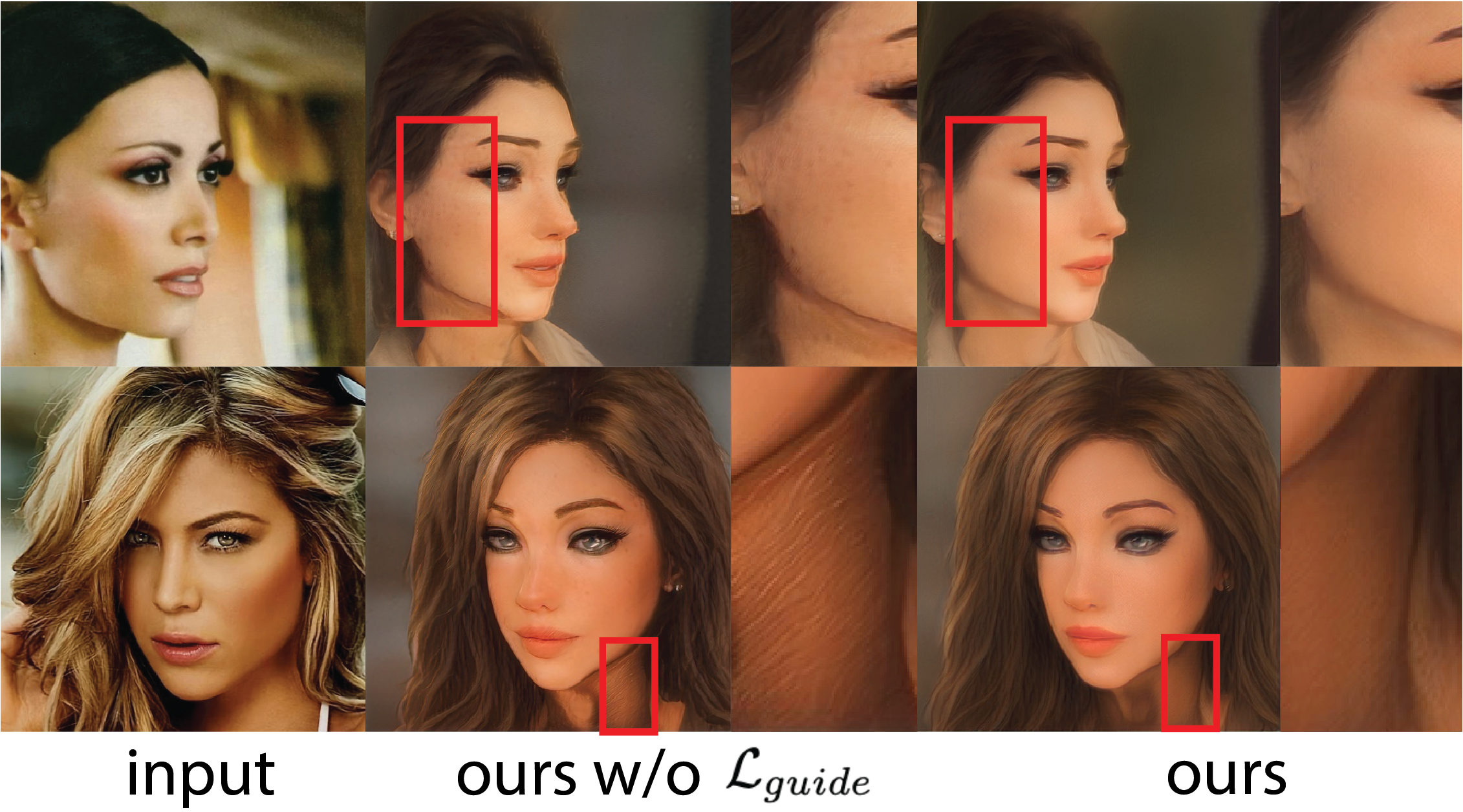}
    \centering
    \caption[Caption for LOF]{Our guided transfer learning loss helps improve generative quality and resolve fine-level visual artifacts.}
    \label{fig:alba_guide}
\end{figure}

To further improve the 3D stylization quality, especially for cases where the inverted codes might be still not well aligned with the original distribution, the stylized images might contain artifacts, such as blurriness. By combining fine-tuning and inversion, we propose a guided transfer learning to enlarge the transfer learning space from its Z space to $W^+$ space with stronger generative stylization capability.

Thanks to the real to stylized face paired data that are produced in \emph{style prior creation} step, we are able to guide the transfer learning of our 3D generator with a reconstruction loss. Given a real image $x$ with estimated camera $p$ and its 2D stylized pair $x_s$, let $\hat{x_s} = \mathcal{G}_{\phi}(\mathcal{E}_{\theta}(x), p)$ we have:
\begin{equation}\label{eq:loss_decoder_rec}
\begin{split}
    \mathcal{L}_{guide} = \mathcal{L}_{2}(x_s, \hat{x_s}) + \mathcal{L}_{lpips}(x_s, \hat{x_s}) 
\end{split}
\end{equation}
This guidance loss can help stabilize the generation training, and also improve the generative quality and user similarity, as illustrated in Fig.~\ref{fig:alba_guide}. 
We fine-tune the 3D generator and discriminator with the encoder and style prior frozen.

\section{Experimental Analysis}

\begin{figure*}[h]
	\centering
	\includegraphics[width=0.95\linewidth]{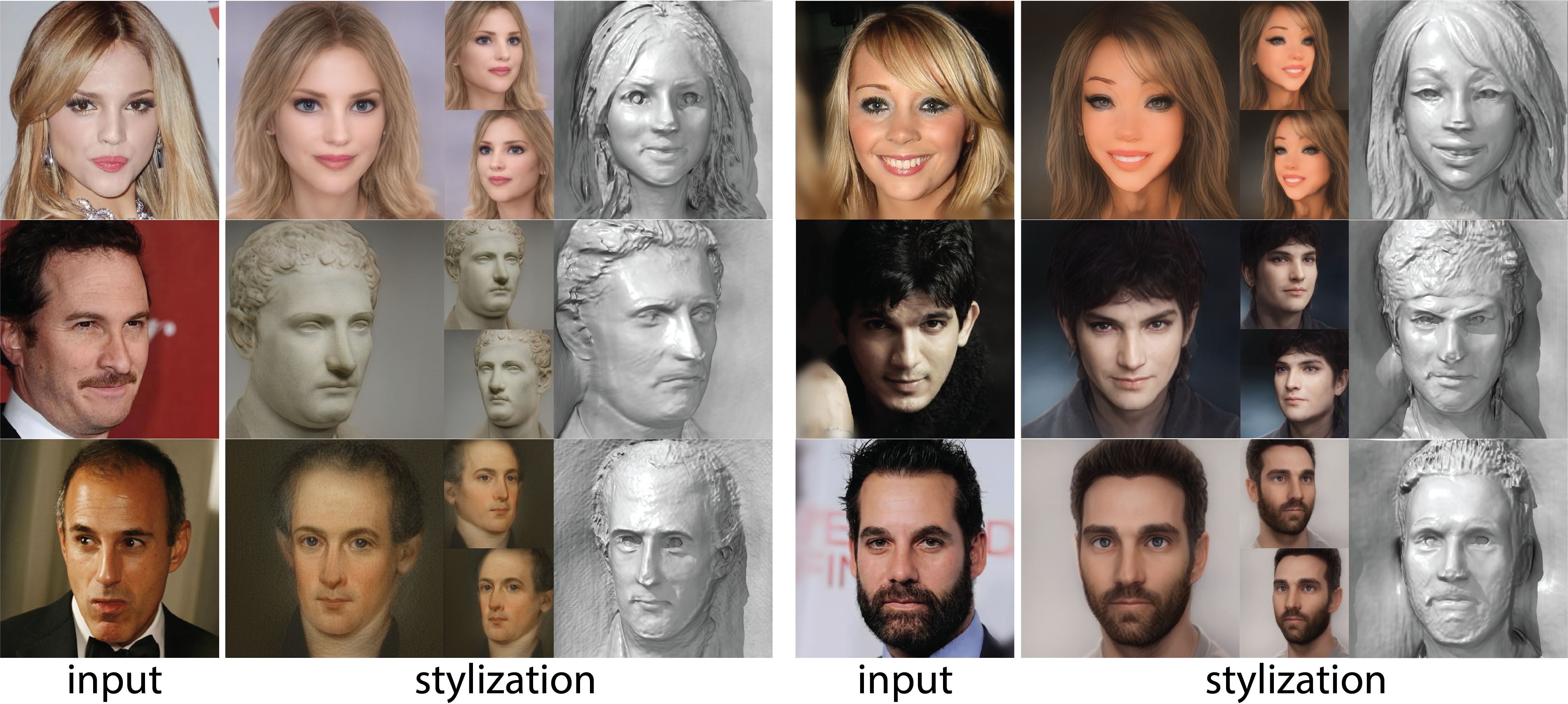} 
	\centering
	\caption[Caption for LOF]{3D artistic portraits generated from a variety of input images. From left to right, we show the input image, and generated stylistic multi-view images and geometry with our pipeline. Please refer to supplementary materials for higher-resolution qualitative results.}
	\label{fig:our_result}
\end{figure*}

\subsection{Implementation Details}
Our encoder is trained on the CelebA-HQ dataset~\cite{CelebAMask-HQ} containing 30,000 high quality face images, where we use the first 28000 for training and the rest 2000 for testing. For consideration of computational efficiency, the input images are down-sampled to 256$\times$256. The pre-trained EG3D uses the weights from the FFHQ 512-128 model \cite{Karras2019stylegan2}.  We empirically set $\lambda_{reg}=0.001$ and $\lambda_{cyc}=1$ with 2 random camera poses. We minimize the objective function for 20 epochs using the Rectified Adam solver \cite{liu2019radam}, with a batch size of 2 and a learning rate of $5 \times 10^{-4}$.

For transfer learning, we collect the initial 20 2D style exemplars from multi-image asset websites\cite{turbosquid, pinterest} for each style, with which we train the style prior. For 3D GAN transfer learning, we use CelebA-HQ as real images in the guided transfer learning loss. Initialized with a pretrained EG3D model, the weights of the generator and discriminators are fine-tuned at a learning rate of 0.002 with a batch size of 4. We limit the number of iterations around 8K images. 


\subsection{Comparisons}
\subsubsection{3D-Aware Stylization}
In Fig.~\ref{fig:our_result}, we present more 3D portrait stylization results from our method. A diverse range of styles demonstrate that our method can robustly handle input images that represent a variety of genders, face shapes and hair styles under different illumination conditions, creating visually appealing, multi-view consistent stylization.

Since there is no prior few-shot 3D stylization methods that we can compare directly, we fine-tune a 3D generator as used in our method, following 2D stylization methods AgileGAN\cite{Song:2021:AgileGAN} and Toonify\cite{pinkney2020resolution}. We name these two hybrid methods as \emph{AgileGAN-EG3D} and \emph{Toonify-EG3D}, and compare to them both quantitatively and qualitatively. 
For \emph{Toonify-EG3D}, we perform direct transfer learning of the generator with the provided style exemplars, and the inversion is achieved with an optimization. For \emph{AgileGAN-EG3D}, in addition to transfer learning the generator, we follow their setting to train a hierarchical variational encoder for image inversion. 

\begin{figure*}
	\centering
	\includegraphics[width=0.95\linewidth]{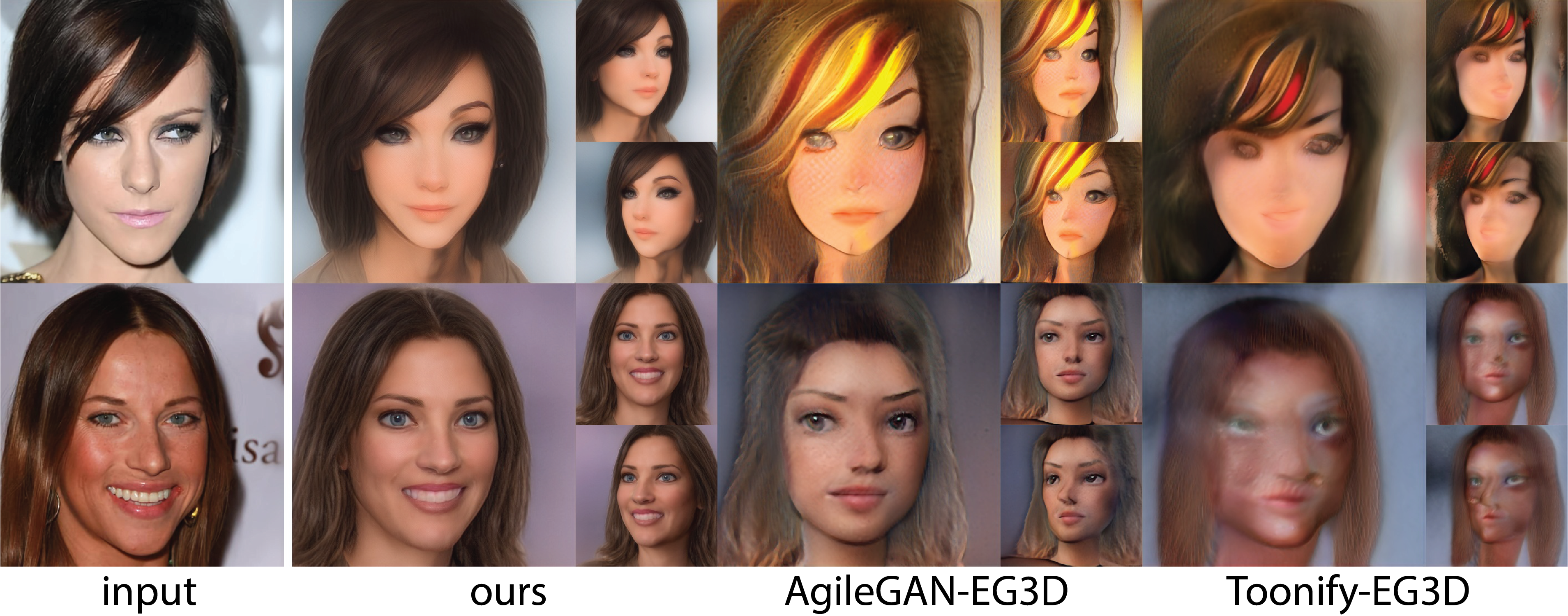}
	\centering
	\caption[Caption for LOF]{Our method visually outperforms direct transfer learning of EG3D generator following 2D few-shot stylization  AgileGAN\cite{Song:2021:AgileGAN} and Toonify\cite{pinkney2020resolution}. Our AgileGAN3D depicts identity-preserved 3D style portraits with fine-level details.}
	\label{fig:compare_sota}
\end{figure*}

\paragraph{Qualitative Evaluation} 
In Fig.~\ref{fig:compare_sota}, we present visual comparisons against the two baseline methods. In constrast with \emph{AgileGAN-EG3D} and \emph{Toonify-EG3D} results exhibiting noticeable artifacts, our approach demonstrates 3D stylization with superior perceptual quality and identity preservation. 

\begin{table}
\caption{Stylization LPIPS  $\downarrow$ for different stylization methods}
\centering
\begin{tabular}{cccc}
    \toprule
       &Ours & AgileGAN-EG3D & Toonify-EG3D \\
    \midrule
    Cartoon &\textbf{0.195} &0.440& 0.481   \\
    Oil Painting &\textbf{0.212} &0.433& 0.525   \\
    Comic &\textbf{0.218} &0.379& 0.486   \\
    Sam Yang &\textbf{0.20} &0.445& 0.506   \\
    Sculpture &\textbf{0.234} &0.469& 0.549   \\
    \bottomrule
\end{tabular}
\label{tab:quantitative}
\end{table}

\paragraph{Quantitative Evaluation}
In  Table~\ref{tab:quantitative}, we also quantitatively measure the visual quality by evaluating a perceptual distance loss between the results of 3D style generator and 2D style prior, which we refer as Stylization LPIPS. The evaluation is performed on CelebA-HQ test images. Given the high quality 2D stylization (without 3D consistent manipulation capability though), we can consider a lower perceptual distance indicating higher visual quality, where our method outperforms the baselines substantially. We also compute a perceptual evaluation with user study, and please refer to our supplemental materials.

\subsubsection{Multiview Manipulation Consistency} 
\begin{figure}
	\centering
	\includegraphics[width=1.0\linewidth]{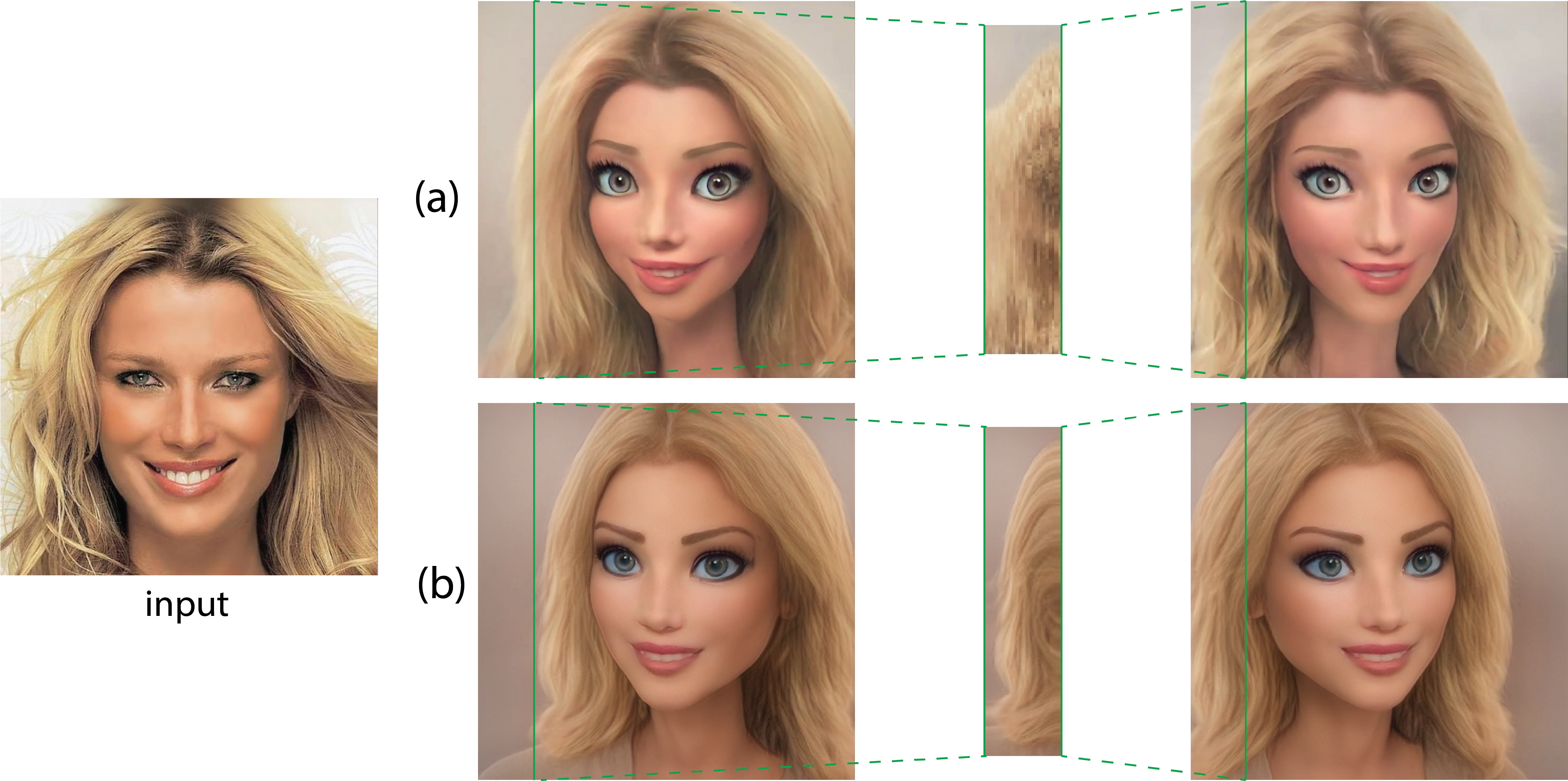}
	\centering
	\caption[Caption for LOF]{
	We compare our method (b) against 2D AgileGAN\cite{Song:2021:AgileGAN} (a) in multiview consistency. We manipulate stylization result from the left to right and horizontally stack a vertical segment of pixels from each generated image (middle). Our method shows a more natural visual transition, indicating better view consistency.}
	\label{fig:smooth_check}
\end{figure}

\begin{figure}
\centering
\includegraphics[width=0.8\linewidth]{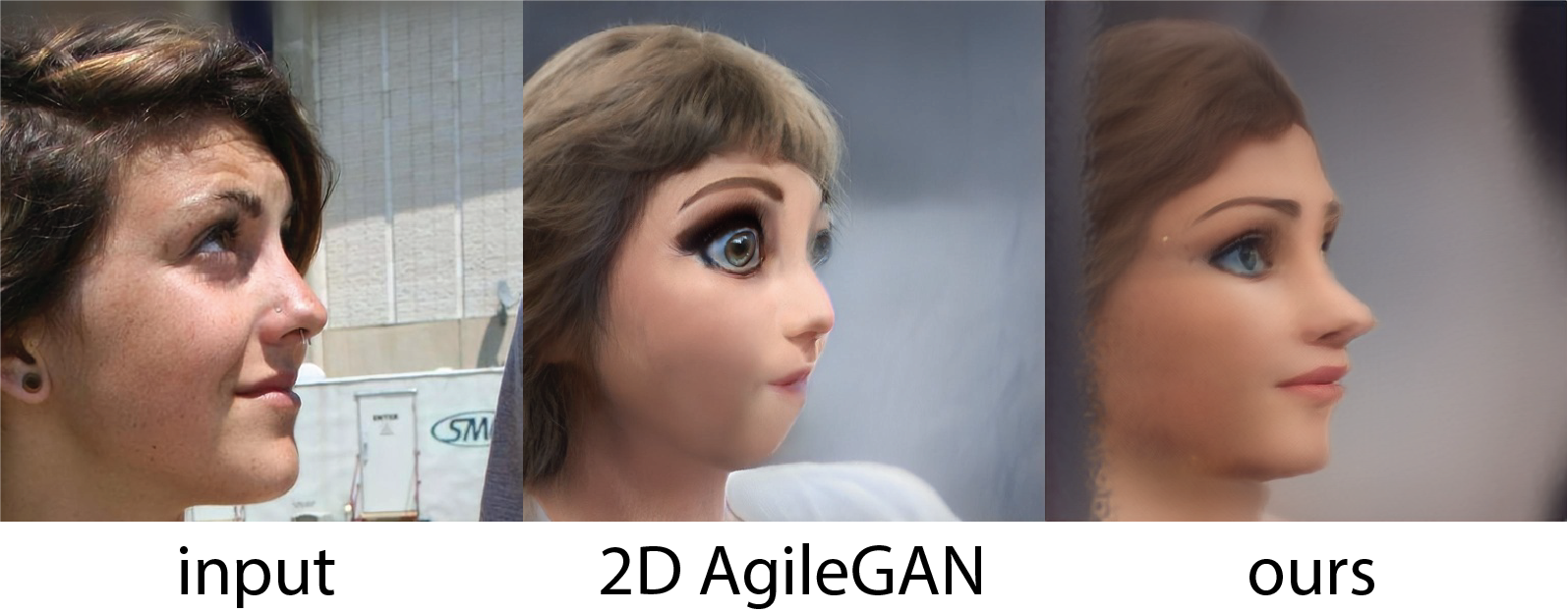}
\centering
\caption[Caption for LOF]{With the prior knowledge of camera extrinsics, our method demonstrates more robust large-pose stylization, compared to 2D AgileGAN\cite{Song:2021:AgileGAN}.}
	\label{fig:large_pose}
\end{figure}

By leveraging a 3D-aware image generator, our method achieves multiview consistent stylization. 2D stylization approaches like AgileGAN\cite{Song:2021:AgileGAN} support limited view manipulation via modifying the latent code\cite{shen2020closedform} but exhibits noticeable visual inconsistency. 
In Fig.~\ref{fig:smooth_check}, we  visually compare the view consistency using Epipolar Line Images (EPI) similar to~\cite{xiang2022gram}, where our method shows smooth and natural pattern transition when rotating the rendering camera.  Additionally benefiting from camera-disentangled image synthesis capability, our AgileGAN3D is more robust in  large-pose stylization as depicted in Fig.~\ref{fig:large_pose}.

\subsection{Ablation Studies}


\begin{figure}
	\centering
	\includegraphics[width=1.0\linewidth]{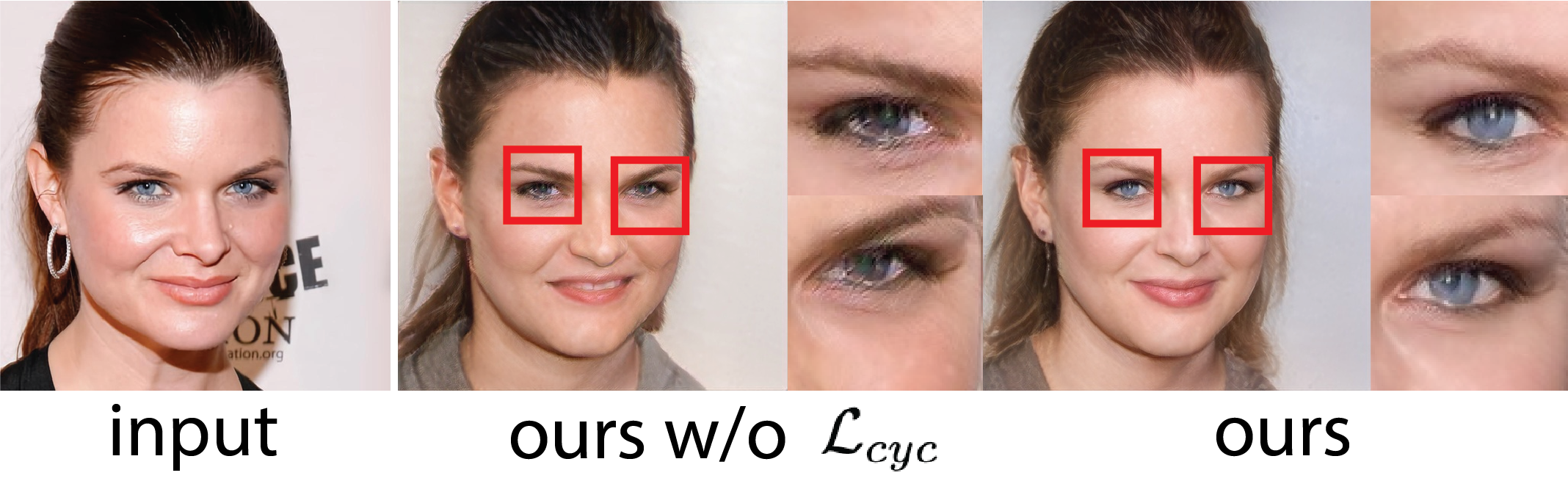}
	\centering
	\caption[Caption for LOF]{Multiview cycle consistent loss improves image inversion with higher perceptual quality and similarity. }
	\label{fig:abla_encoder}
\end{figure}

\paragraph{Inversion Learning}
In Fig.~\ref{fig:abla_encoder}, we evaluate the efficacy of our cycle consistency loss introduced in our encoder for image inversion. With our loss, the encoder presents higher perceptual quality and identity similarity, numerically evidenced with better reconstruction losses in Tab.~\ref{tab:encoder_quantitative}. 

\begin{table}
	\caption{Quantitative ablation of cycle consistency loss, evaluated from 2K testing images.}
	\centering
	\begin{tabular}{cccc}
		\toprule
		Algorithm & MSE  $\downarrow$ & SSIM $\uparrow$ & LPIPS $\downarrow$ \\
		\midrule
		Ours w/o Cycle Loss & 0.0203 & 0.525 & 0.317 \\
		Ours  & \textbf{0.0200} & \textbf{0.543} &\textbf{0.194} \\
		\bottomrule
	\end{tabular}
	\label{tab:encoder_quantitative}
\end{table}

\paragraph{Transfer Learning}
We evaluate the effect of training sample quantity over the stylization quality both numerically in Tab.~\ref{tab:transfer_quantitative} and qualitatively in Fig.~\ref{fig:alba_volume}. The experiment is performed over sam yang style. We compare against 2D AgileGAN stylization over test CelebA-HD images, where we observe closer perceptual quality as the number of training exemplars increase, demonstrating the efficacy of our augmented transfer learning. Additionally, our guided transfer learning further improves the perceptual score,  as also visually evidenced in Fig.~\ref{fig:alba_guide}. We note that using camera labels estimated from style images leads to degenerated perceptual quality. That being said, our paired camera labels from real images help 3D GAN transfer learning. 


\begin{table}
\caption{Stylization Perceptual scores $\downarrow$ with different training samples}
\centering
\begin{tabular}{cc}
\toprule
	\# Training samples & LPIPS$\downarrow$ \\
		\midrule
    20 &0.41   \\
    100 &0.252 \\
    1000 &0.236    \\
    8000 &0.227    \\
    \midrule
    8000(with guided loss, without accurate pose) &0.211 \\
    8000(with guided loss) &\textbf{0.200} \\
\bottomrule
\end{tabular}
\label{tab:transfer_quantitative}
\end{table}







\section{Conclusion}
We presented \emph{AgileGAN3D}, the first few-shot pipeline generating high quality 3D stylistic portraits with detailed in-style geometry from a single user image, which sheds light on many potential applications. Our method only uses a limited number (around 20) of unpaired 2D style exemplars for a new target style. This is achieved via a novel framework incorporating \emph{style prior creation} into \emph{guided transfer learning}, which addresses the inadequate supervision issue of 3D GAN transfer learning with accurate camera labels. We also introduce a 3D GAN inversion module with \emph{multi-view consistency loss} to improve identity preservation while achieving appealing stylization quality.
Experimental results show that the algorithm produces high-quality multi-view consistent stylized 3D portraits.

\begin{figure}
	\centering
	\includegraphics[width=1.0\linewidth]{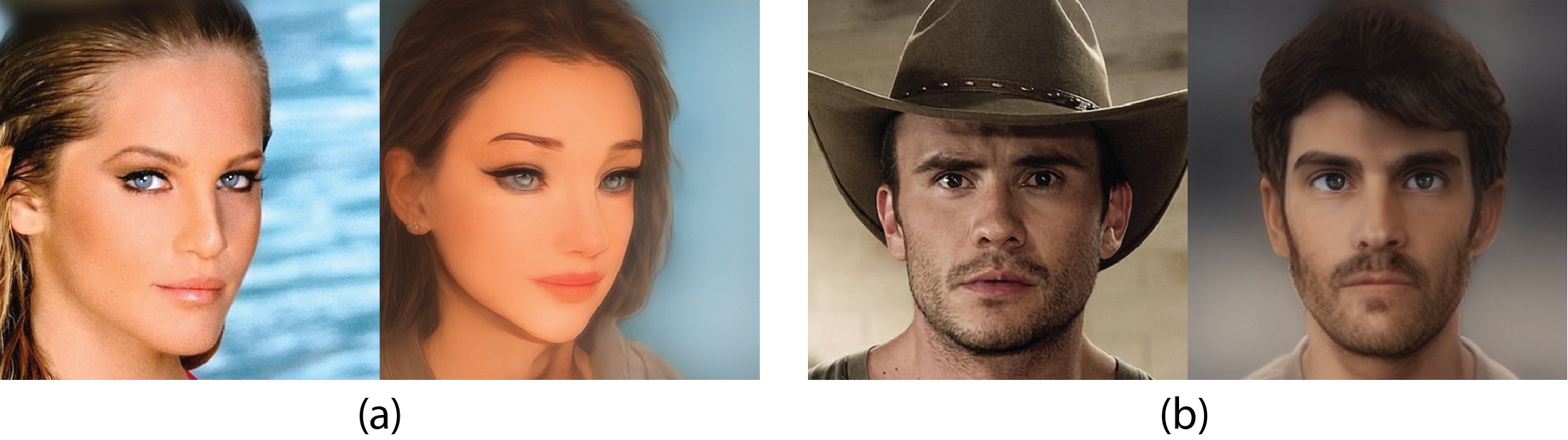}
	\centering
	\caption[Caption for LOF]{Failure examples. (a) inconsistent gaze directions, (b) unmodeled hat. }
	\label{fig:limitation}
\end{figure}

\paragraph{Limitations} We presented a variety of compelling 3D portrait stylization results, but there is still space for further improvement in our framework.  Fig.~\ref{fig:limitation} shows two example failure cases. (a) In some situations, we found that the generated eye gaze direction is biased towards frontal gaze, which may not be consistent with the input. (b) Occasionally, our approach fail to preserve accessories such as hat and glasses after stylization, as such cases are under-represented in the input datasets. These problems could potentially be mitigated by including more diverse input exemplars.

\paragraph{Societal Impact} Our work focuses on improving stylization quality of 3D GANs in technical aspects and is not designed for any malicious uses. However, inappropriate usage of the proposed method might generate undesired fake images, which is a well-known issue of portrait stylization algorithms.


{\small
\bibliographystyle{ieee_fullname}
\bibliography{egbib}
}

\end{document}